\documentclass[conference]{IEEEtran}
\IEEEoverridecommandlockouts
\usepackage{cite}
\usepackage{amsmath,amssymb,amsfonts}
\usepackage{algorithmic}
\usepackage{graphicx}
\usepackage{textcomp}
\usepackage{xcolor}
\def\BibTeX{{\rm B\kern-.05em{\sc i\kern-.025em b}\kern-.08em
    T\kern-.1667em\lower.7ex\hbox{E}\kern-.125emX}}

\newtheorem{theorem}{\it Theorem}
\newtheorem{lemma}{\it Lemma}

\newtheorem{corollary}{\it Corollary}
\newtheorem{proposition}{\it Proposition}

\DeclareMathOperator*{\esssup}{ess\,sup}

\begin{document}

\title{Fundamental Limitations in Sequential Prediction and Recursive Algorithms: $\mathcal{L}_{p}$ Bounds via an Entropic Analysis\\
%{\footnotesize \textsuperscript{*}Note: Sub-titles are not captured in Xplore and
%should not be used}
\thanks{This work was supported in part by NSF under grant ECCS-1847056 and SES-1541164, in part by a U. S. DOT grant through C2SMART Center at NYU, and in part by the U.S. DHS through the CIRI under Grant 2015-ST-061-CIRC01.}
}

%Fundamental Limitations in Sequential Prediction: Entropic Analysis, $\mathcal{L}_{p}$ Bounds, and Beyond

\author{\IEEEauthorblockN{Song Fang and Quanyan Zhu}
\IEEEauthorblockA{\textit{Department of Electrical and Computer Engineering, New York University, New York, USA}\\
{\tt \small song.fang@nyu.edu, quanyan.zhu@nyu.edu}
}
%\and
%\IEEEauthorblockN{2\textsuperscript{nd} Given Name Surname}
%\IEEEauthorblockA{\textit{dept. name of organization (of Aff.)} \\
%\textit{name of organization (of Aff.)}\\
%City, Country \\
%email address or ORCID}
%\and
%\IEEEauthorblockN{3\textsuperscript{rd} Given Name Surname}
%\IEEEauthorblockA{\textit{dept. name of organization (of Aff.)} \\
%\textit{name of organization (of Aff.)}\\
%City, Country \\
%email address or ORCID}
%\and
%\IEEEauthorblockN{4\textsuperscript{th} Given Name Surname}
%\IEEEauthorblockA{\textit{dept. name of organization (of Aff.)} \\
%\textit{name of organization (of Aff.)}\\
%City, Country \\
%email address or ORCID}
%\and
%\IEEEauthorblockN{5\textsuperscript{th} Given Name Surname}
%\IEEEauthorblockA{\textit{dept. name of organization (of Aff.)} \\
%\textit{name of organization (of Aff.)}\\
%City, Country \\
%email address or ORCID}
%\and
%\IEEEauthorblockN{6\textsuperscript{th} Given Name Surname}
%\IEEEauthorblockA{\textit{dept. name of organization (of Aff.)} \\
%\textit{name of organization (of Aff.)}\\
%City, Country \\
%email address or ORCID}
}

\maketitle

\begin{abstract}
In this paper, we obtain fundamental $\mathcal{L}_{p}$ bounds in sequential prediction and recursive algorithms via an entropic analysis. Both of the two classes of problems are examined by investigating the underlying entropic relationships of the data involved, and the coresponding bounds are all shown to depend merely on the conditional entropy relations therein.
\end{abstract}


\section{Introduction}
\label{sec:intro}

%%Nowadays machine learning techniques are becoming more and more prevalent in real-time systems such as real-time signal processing, feedback control, and robotics systems. In such systems, on one hand, decisions on the actions are to be made in a sequential manner (sequential decision making); on the other hand, dynamics of the systems as well as the environment that are determined by physical laws will play an indispensable role and must be taken into consideration (interaction with real world). In this trend, it is becoming more and more critical to be fully aware of the performance limits of the sequential machine learning algorithms that are to be embedded in the autonomous systems operating in real world, especially in scenarios where worst-case performance guarantees are required and must be strictly imposed. 
%%
%%Sequential prediction (or sequence prediction) \cite{sun2001sequence, dietterich2002machine, cesa2006prediction, shalev2012online, rakhlin2014statistical} has been an important component of sequential learning. 
%%
%%In this paper, we will utilize information theory to analyze the fundamental performance bounds of sequential prediction. 
%%
%%Extrapolation
%%
%%Innovation
%%
%%Estimation Prediction

Information theory \cite{Cov:06} was originally developed to analyze the fundamental limitations in communication, which may represent any systems that involve information transmission from one point to another point, or, as Shannon put it \cite{shannon1998mathematical}, any systems that involve ``reproducing at one point either exactly or approximately a message selected at another point". In a broad sense, the machine learning approaches may be viewed as information transmission (or, message reproducing) processes, as if extracting as much ``information" as possible out of the training data (cf. discussions in, e.g., \cite{mackay2003information, tishby2015deep, shwartz2017opening}) and then transmitting the information to the test data, so as to reduce as much as possible the ``uncertainty" or ``randomness" contained in the latter. In sequential prediction, this ``information extraction $\to$ information transmission $\to$ uncertainty reduction" process is done in a sequential manner, while in recursive algorithms, the process is carried out in a recursive way. By virtue of this analogy in terms of ``message reproducing", in this paper we examine the fundamental limitations in sequential prediction and recursive algorithms via an information-theoretic approach, that is, via an entropic analysis.

In linear prediction theory \cite{kailath1974view, makhoul1975linear, pourahmadi2001foundations, vaidyanathan2007theory, caines2018linear}, which has been an important branch of signal processing, the Kolmogorov--Szeg\"o formula \cite{Pap:02, vaidyanathan2007theory, lindquist2015linear, chen2018role, FangACC18, FangCDC17} provides a fundamental bound on the variance of prediction error for the linear prediction of Gaussian sequences. In this paper, we go beyond the linear Gaussian case; instead, we consider the generic sequential prediction setting in which the sequence to be predicted can be with arbitrary distributions while the predictor can be any arbitrarily causal, and we derive the fundamental $\mathcal{L}_{p}$ bounds (more general than the variance bounds, which are essentially $\mathcal{L}_{2}$ bounds, i.e., the special case of $p=2$) on the prediction errors. In particular, we derive the prediction bounds by investigating the underlying entropic relationships of the data points composing the sequences, while the obtained bounds can be characterized explicitly by the conditional entropy of the data point to be predicted given the previous data points. Similarly, we study the fundamental $\mathcal{L}_{p}$ bounds on the recursive differences in recursive algortithms, and it is seen that the recursive difference bounds can be quantified explicitly by the conditional entropy of current noise conditioned on the past noises and the initial state of the recursive algorithm.

The remainder of the paper is organized as follows. Section~II introduces the technical preliminaries. In Section~III, we introduce the fundamental $\mathcal{L}_{p}$ bounds in sequential prediction. Section~IV presents the fundamental $\mathcal{L}_{p}$ bounds in recursive algorithms. Concluding remarks are given in Section~V.

Note that special cases (for $p=2$ and $p=\infty$; see Section~III) of the prediction bounds have been presented in our previous papers \cite{FangITW19} and \cite{FangMLSP19}, while this paper provides a unifying framework for $p \geq 1$. On the other hand, the recursive algorithm bounds (see Section~IV) did not appear in \cite{FangITW19} or \cite{FangMLSP19}, although the special cases for $p=2$ and $p=\infty$ have been included without proofs in their updated arXiv versions \cite{FangITW19arxiv} and \cite{fang2019generic}. Note also that due to lack of space, discussions on the relation to the Kolmogorov--Szeg\"o formula are not included and may be found in \cite{FangCISS20arxiv}.

\section{Preliminaries}

In this paper, we consider real-valued continuous random variables and vectors, as well as discrete-time stochastic processes they compose. All random variables, random vectors, and stochastic processes are assumed to be zero-mean for simplicity and without loss of generality. We represent random variables and vectors using boldface letters. Given a stochastic process $\left\{ \mathbf{x}_{k}\right\}, \mathbf{x}_{k} \in \mathbb{R}$, we denote the sequence $\mathbf{x}_0,\ldots,\mathbf{x}_{k}$ by the random vector $\mathbf{x}_{0,\ldots,k}=\left[\mathbf{x}_0~\cdots~\mathbf{x}_{k}\right]^T$. The logarithm is defined with base $2$. All functions are assumed to be measurable. 
A stochastic process $\left\{ \mathbf{x}_{k}\right\}$ is said to be asymptotically stationary if it is stationary as $k \to \infty$, and herein stationarity means strict stationarity unless otherwise specified \cite{Pap:02}. In addition, a process being asymptotically stationary implies that it is asymptotically mean stationary \cite{gray2011entropy}.

Definitions and properties of the information-theoretic notions that will be used in this paper, including differential entropy $h\left( \mathbf{x} \right)$, conditional entropy $
h\left(\mathbf{x} \middle| \mathbf{y}\right)$, entropy rate $h_\infty \left(\mathbf{x}\right)$, and mutual information $I\left(\mathbf{x};\mathbf{y}\right)$, can be found in, e.g., \cite{Cov:06}.

On the other hand, the next lemma \cite{dolinar1991maximum} presents maximum-entropy probability distributions under $\mathcal{L}_{p}$-norm constraints.

\begin{lemma} \label{maximum}
	Consider a random variable $\mathbf{x} \in \mathbb{R}$ with $\mathcal{L}_{p}$ norm $\left[ \mathbb{E} \left( \left| \mathbf{x} \right|^{p} \right) \right]^{\frac{1}{p}} = \mu,~p \geq 1$.
	Then,  
	\begin{flalign} 
	h \left( \mathbf{x} \right) 
	\leq \log \left[ 2 \Gamma \left( \frac{p+1}{p} \right) \left( p \mathrm{e} \right)^{\frac{1}{p}} \mu \right], \nonumber
	\end{flalign}
	%	or equivalently, 
	%	\begin{flalign} 
	%	\left| \mathrm{supp} \left( \mathbf{x} \right) \right|
	%	\geq 2^{h \left( \mathbf{x} \right)}, \nonumber
	%	\end{flalign}
	where equality holds if and only if $\mathbf{x}$ is with probability density
	\begin{flalign}
	f_{\mathbf{x}} \left( x \right)
	= \frac{ \mathrm{e}^{- \left| x \right|^{p} / \left( p \mu^{p} \right)} }{2 \Gamma \left( \frac{p+1}{p} \right) p^{\frac{1}{p}} \mu}. \nonumber
	\end{flalign}
	Herein, $\Gamma \left( \cdot \right)$ denotes the Gamma function.
\end{lemma}

In particular, when $p \to \infty$, 
\begin{flalign}
\lim_{p \to \infty} \left[ \mathbb{E} \left( \left| \mathbf{x} \right|^{p} \right) \right]^{\frac{1}{p}} = \esssup_{ f_{\mathbf{x}} \left( x \right) > 0} \left| \mathbf{x} \right|, \nonumber
\end{flalign}
and
\begin{flalign} 
\lim_{p \to \infty} \log \left[ 2 \Gamma \left( \frac{p+1}{p} \right) \left( p \mathrm{e} \right)^{\frac{1}{p}} \mu \right] = \log \left( 2 \mu \right), \nonumber
\end{flalign}
while
\begin{flalign}
\lim_{p \to \infty}\frac{ \mathrm{e}^{- \left| x \right|^{p} / \left( p \mu^{p} \right)} }{2 \Gamma \left( \frac{p+1}{p} \right) p^{\frac{1}{p}} \mu}
= 
\left\{ \begin{array}{cc}
\frac{1}{2 \mu}, & \left| x \right| \leq \mu,\\
0, & \left| x \right| > \mu,
\end{array} \right. \nonumber
\end{flalign}

In fact, an alternative form of Lemma~\ref{maximum} can be obtained as follows.

\begin{proposition}
	Consider a random variable $\mathbf{x} \in \mathbb{R}$ with entropy $h \left(  \mathbf{x} \right)$.
	Then,  
	\begin{flalign} \label{alternative}
	\left[ \mathbb{E} \left( \left| \mathbf{x} \right|^{p} \right) \right]^{\frac{1}{p}}
	\geq \frac{2^{h \left(  \mathbf{x} \right)}}{2 \Gamma \left( \frac{p+1}{p} \right) \left( p \mathrm{e} \right)^{\frac{1}{p}}},
	\end{flalign}
	where equality holds if and only if $\mathbf{x}$ is with probability density
	\begin{flalign}
	f_{\mathbf{x}} \left( x \right)
	= \frac{ \mathrm{e}^{- \left| x \right|^{p} / \left( p \mu^{p} \right)} }{2 \Gamma \left( \frac{p+1}{p} \right) p^{\frac{1}{p}} \mu}. 
	\end{flalign}	
	Herein, $\mu$ is a normalizing factor.
\end{proposition}

As a matter of fact, when equality is achieved in \eqref{alternative}, it can be shown that
\begin{flalign}
\mu 
= \frac{2^{h \left( \mathbf{x} \right)}}{2 \Gamma \left( \frac{p+1}{p} \right) \left( p \mathrm{e} \right)^{\frac{1}{p}}}
= \left[ \mathbb{E} \left( \left| \mathbf{x} \right|^{p} \right) \right]^{\frac{1}{p}}.
\end{flalign}

In particular, when $p \to \infty$, \eqref{alternative} reduces to
\begin{flalign}
\esssup_{ f_{\mathbf{x}} \left( x \right) > 0} \left| \mathbf{x} \right| \geq \frac{2^{h \left(  \mathbf{x} \right)}}{2},
\end{flalign}
where equality holds if and only if $\mathbf{x}$ is with probability density
\begin{flalign}
f_{\mathbf{x}} \left( x \right)
= 
\left\{ \begin{array}{cc}
\frac{1}{2 \mu}, & \left| x \right| \leq \mu,\\
0, & \left| x \right| > \mu,
\end{array} \right.
\end{flalign}
that is to say, if and only if $\mathbf{x}$ is uniform, with 
\begin{flalign}
\mu = \frac{2^{h \left(  \mathbf{x} \right)}}{2} = \esssup_{ f_{\mathbf{x}} \left( x \right) > 0} \left| \mathbf{x} \right|.
\end{flalign}

\section{$\mathcal{L}_{p}$ Bounds in Sequential Prediction}

In what follows, we provide a generic bound on the $\mathcal{L}_{p}$ norm of the prediction error for when predicting a data point sequentially based on its previous data points. 

\begin{theorem} \label{MIMOprediction}
	Consider a stochastic process $\left\{ \mathbf{x}_{k} \right\}, \mathbf{x}_{k} \in \mathbb{R}$. Denote the $1$-step ahead prediction (in the rest of the paper, ``$1$-step ahead prediction" will be abbreviated as ``$1$-step prediction" for simplicity) of $\mathbf{x}_{k}$ by $\widehat{\mathbf{x}}_{k} = g_{k} \left( \mathbf{x}_{0,\ldots,k-1} \right)$.
	Then,  
	\begin{flalign} \label{MIMOprediction1}
	\left[ \mathbb{E} \left( \left| \mathbf{x}_{k} - \widehat{\mathbf{x}}_{k} \right|^{p} \right) \right]^{\frac{1}{p}}
	\geq \frac{2^{h \left( \mathbf{x}_k | \mathbf{x}_{0,\ldots,k-1} \right)}}{2 \Gamma \left( \frac{p+1}{p} \right) \left( p \mathrm{e} \right)^{\frac{1}{p}}},
	\end{flalign}
	where equality holds if and only if $\mathbf{x}_{k} - \widehat{\mathbf{x}}_{k}$ is with probability density 
	\begin{flalign} \label{distribution}
	f_{\mathbf{x}_{k} - \widehat{\mathbf{x}}_{k}} \left( x \right)
	= \frac{ \mathrm{e}^{- \left| x \right|^{p} / \left( p \mu^{p} \right)} }{2 \Gamma \left( \frac{p+1}{p} \right) p^{\frac{1}{p}} \mu},
	\end{flalign}
	and $I \left( \mathbf{x}_{k} - \widehat{\mathbf{x}}_{k}; \mathbf{x}_{0,\ldots,k-1} \right) = 0$.
\end{theorem}

{\it Proof.}
It is known from Lemma~\ref{maximum} that 
\begin{flalign} 
\left[ \mathbb{E} \left( \left| \mathbf{x}_{k} - \widehat{\mathbf{x}}_{k} \right|^{p} \right) \right]^{\frac{1}{p}}
\geq \frac{2^{h \left(  \mathbf{x}_{k} - \widehat{\mathbf{x}}_{k} \right)}}{2 \Gamma \left( \frac{p+1}{p} \right) \left( p \mathrm{e} \right)^{\frac{1}{p}}}, \nonumber
\end{flalign} 
where equality holds if and only if $\mathbf{x}_{k} - \widehat{\mathbf{x}}_{k}$ is with probability density \eqref{distribution}. Meanwhile,
\begin{flalign} 
&h \left( \mathbf{x}_{k} - \widehat{\mathbf{x}}_{k} \right) 
= h \left( \mathbf{x}_{k} - \widehat{\mathbf{x}}_{k} |  \mathbf{x}_{0,\ldots,k-1} \right)  + I \left( \mathbf{x}_{k} - \widehat{\mathbf{x}}_{k}; \mathbf{x}_{0,\ldots,k-1} \right) \nonumber \\
& = h \left( \mathbf{x}_k - g_{k} \left(  \mathbf{x}_{0,\ldots,k-1} \right) |  \mathbf{x}_{0,\ldots,k-1} \right)  + I \left( \mathbf{x}_{k} - \widehat{\mathbf{x}}_{k}; \mathbf{x}_{0,\ldots,k-1} \right) \nonumber \\
& = h \left( \mathbf{x}_k | \mathbf{x}_{0,\ldots,k-1} \right) + I \left( \mathbf{x}_{k} - \widehat{\mathbf{x}}_{k}; \mathbf{x}_{0,\ldots,k-1} \right) \nonumber \\
& \geq h \left( \mathbf{x}_k | \mathbf{x}_{0,\ldots,k-1} \right). \nonumber
\end{flalign}
As a result,
$
2^{ h \left( \mathbf{x}_{k} - \widehat{\mathbf{x}}_{k} \right)} 
\geq 2^{h \left( \mathbf{x}_k | \mathbf{x}_{0,\ldots,k-1} \right)},
$
where equality holds if and only if $I \left( \mathbf{x}_{k} - \widehat{\mathbf{x}}_{k}; \mathbf{x}_{0,\ldots,k-1} \right) = 0$.
Therefore,
\begin{flalign}
\left[ \mathbb{E} \left( \left| \mathbf{x}_{k} - \widehat{\mathbf{x}}_{k} \right|^{p} \right) \right]^{\frac{1}{p}}
\geq \frac{2^{h \left( \mathbf{x}_k | \mathbf{x}_{0,\ldots,k-1} \right)}}{2 \Gamma \left( \frac{p+1}{p} \right) \left( p \mathrm{e} \right)^{\frac{1}{p}}}, \nonumber
\end{flalign}
where equality holds if and only if $\mathbf{x}_{k} - \widehat{\mathbf{x}}_{k}$ is with probability density \eqref{distribution} and $I \left( \mathbf{x}_{k} - \widehat{\mathbf{x}}_{k}; \mathbf{x}_{0,\ldots,k-1} \right) = 0$.
\hfill$\square$

Again, $\mu$ is a normalizing factor herein. In addition, when equality is achieved in \eqref{MIMOprediction1}, it can be shown that
\begin{flalign}
\mu 
= \frac{2^{h \left( \mathbf{x}_k | \mathbf{x}_{0,\ldots,k-1} \right)}}{2 \Gamma \left( \frac{p+1}{p} \right) \left( p \mathrm{e} \right)^{\frac{1}{p}}}
= \left[ \mathbb{E} \left( \left| \mathbf{x}_{k} - \widehat{\mathbf{x}}_{k} \right|^{p} \right) \right]^{\frac{1}{p}}.
\end{flalign}
Note that for the rest of the paper, $\mu$ will always be a normalizing factor as of here, and its value can always be determined in a similar manner as well. Hence, for simplicity, we may skip dicussions concerning how to decide $\mu$ in the rest of the paper.

On the other hand, it is seen that the prediction bound depends only on the conditional entropy of the data point $\mathbf{x}_{k}$ to be predicted given the previous data points $\mathbf{x}_{0,\ldots,k-1}$, i.e.,, the amount of ``randomness" contained in $\mathbf{x}_{k}$ given $\mathbf{x}_{0,\ldots,k-1}$. As such, if $\mathbf{x}_{0,\ldots,k-1}$ provide more/less information of $\mathbf{x}_{k}$, then the conditional entropy becomes smaller/larger, and thus the bound becomes smaller/larger. 
%In particular, if $\left\{ \mathbf{x}_{k} \right\}$ is a Markov process \cite{Cov:06}, then $h \left( \mathbf{x}_k | \mathbf{x}_{0,\ldots,k-1} \right) = h \left( \mathbf{x}_k | \mathbf{x}_{k-1} \right)$, and hence \eqref{MIMOprediction1} reduces to
%\begin{flalign}
%\left| \mathrm{supp} \left( \mathbf{x}_{k} - \widehat{\mathbf{x}}_{k} \right) \right|
%\geq 2^{ h \left( \mathbf{x}_k | \mathbf{x}_{k-1} \right)},
%\end{flalign}
%where equality holds if and only if $\mathbf{x}_{k} - \widehat{\mathbf{x}}_{k}$ is uniform and $I \left( \mathbf{x}_{k} - \widehat{\mathbf{x}}_{k}; \mathbf{x}_{k-1} \right) = 0$. 
%In the limit when $\mathbf{x}_{k}$ is independent of $\mathbf{x}_{0,\ldots,k-1}$, we have $h \left( \mathbf{x}_k | \mathbf{x}_{0,\ldots,k-1} \right) = h \left( \mathbf{x}_k \right)$, and thus
%$
%\left| \mathrm{supp} \left( \mathbf{x}_{k} - \widehat{\mathbf{x}}_{k} \right) \right|
%\geq 2^{h \left( \mathbf{x}_k \right)}. 
%$

In addition, equality in \eqref{MIMOprediction1} holds if and only if the innovation $\mathbf{x}_{k} - \widehat{\mathbf{x}}_{k} $ is with probability \eqref{distribution}, and contains no information of the previous data points $\mathbf{x}_{0,\ldots,k-1}$; it is as if all the ``information" that may be utilized to reduce the prediction error's $\mathcal{L}_{p}$ norm has been extracted.

\subsection{Special Cases} \label{special}

We now consider the special cases of when $p=1$, $p=2$, and $p=\infty$, respectively.

\subsubsection{When $p=1$} The next corollary follows.

\begin{corollary}
	Consider a stochastic process $\left\{ \mathbf{x}_{k} \right\}, \mathbf{x}_{k} \in \mathbb{R}$. Denote the $1$-step prediction of $\mathbf{x}_{k}$ by $\widehat{\mathbf{x}}_{k} = g_{k} \left( \mathbf{x}_{0,\ldots,k-1} \right)$.
	Then,  
	\begin{flalign} 
	\mathbb{E} \left| \mathbf{x}_{k} - \widehat{\mathbf{x}}_{k} \right|
	\geq \frac{2^{h \left( \mathbf{x}_k | \mathbf{x}_{0,\ldots,k-1} \right)}}{2\mathrm{e}},
	\end{flalign}
	where equality holds if and only if $\mathbf{x}_{k} - \widehat{\mathbf{x}}_{k}$ is with probability density 
	\begin{flalign}
	f_{\mathbf{x}_{k} - \widehat{\mathbf{x}}_{k}} \left( x \right)
	= \frac{ \mathrm{e}^{- \left| x \right| / \mu } }{2 \mu},
	\end{flalign}
	that is to say, if and only if $\mathbf{x}_{k} - \widehat{\mathbf{x}}_{k}$ is Laplace,
	and $I \left( \mathbf{x}_{k} - \widehat{\mathbf{x}}_{k}; \mathbf{x}_{0,\ldots,k-1} \right) = 0$.
\end{corollary}

\subsubsection{When $p=2$} The next corollary follows.

\begin{corollary}
    Consider a stochastic process $\left\{ \mathbf{x}_{k} \right\}, \mathbf{x}_{k} \in \mathbb{R}$. Denote the $1$-step prediction of $\mathbf{x}_{k}$ by $\widehat{\mathbf{x}}_{k} = g_{k} \left( \mathbf{x}_{0,\ldots,k-1} \right)$.
	Then,  
	\begin{flalign} \label{variance}
	\left\{ \mathbb{E} \left[ \left( \mathbf{x}_{k} - \widehat{\mathbf{x}}_{k} \right)^{2} \right]  \right\}^{\frac{1}{2}}
	\geq \frac{2^{h \left( \mathbf{x}_k | \mathbf{x}_{0,\ldots,k-1} \right)}}{\left( 2 \pi \mathrm{e} \right)^{\frac{1}{2}}},
	\end{flalign}
	where equality holds if and only if $\mathbf{x}_{k} - \widehat{\mathbf{x}}_{k}$ is with probability density 
	\begin{flalign}
	f_{\mathbf{x}_{k} - \widehat{\mathbf{x}}_{k}} \left( x \right)
	= \frac{ \mathrm{e}^{- x^{2} / \left( 2 \mu^{2} \right)} }{ \left(2 \pi \mu^2 \right)^{\frac{1}{2}} },
	\end{flalign}
	that is to say, if and only if $\mathbf{x}_{k} - \widehat{\mathbf{x}}_{k}$ is Gaussian,
	and $I \left( \mathbf{x}_{k} - \widehat{\mathbf{x}}_{k}; \mathbf{x}_{0,\ldots,k-1} \right) = 0$.
\end{corollary}

It is clear that \eqref{variance} may be rewritten as
\begin{flalign} 
\mathbb{E} \left[ \left( \mathbf{x}_{k} - \widehat{\mathbf{x}}_{k} \right)^{2} \right]
\geq \frac{2^{2 h \left( \mathbf{x}_k | \mathbf{x}_{0,\ldots,k-1} \right)}}{ 2 \pi \mathrm{e}},
\end{flalign}
which conincides with the conclusions in \cite{Cov:06} and \cite{FangITW19}.

\subsubsection{When $p=\infty$} The next corollary follows.

\begin{corollary} 
	Consider a stochastic process $\left\{ \mathbf{x}_{k} \right\}, \mathbf{x}_{k} \in \mathbb{R}$. Denote the $1$-step prediction of $\mathbf{x}_{k}$ by $\widehat{\mathbf{x}}_{k} = g_{k} \left( \mathbf{x}_{0,\ldots,k-1} \right)$.
	Then,  
	\begin{flalign} \label{MD}
	\esssup_{ f_{\mathbf{x}_{k} - \widehat{\mathbf{x}}_{k}} \left( x \right) > 0} \left| \mathbf{x}_{k} - \widehat{\mathbf{x}}_{k} \right|
	\geq \frac{2^{h \left( \mathbf{x}_k | \mathbf{x}_{0,\ldots,k-1} \right)}}{2},
	\end{flalign}
	where equality holds if and only if $\mathbf{x}_{k} - \widehat{\mathbf{x}}_{k}$ is with probability density 
	\begin{flalign}
	f_{\mathbf{x}_{k} - \widehat{\mathbf{x}}_{k}} \left( x \right)
	= \left\{ \begin{array}{cc}
	\frac{1}{2 \mu}, & \left| x \right| \leq \mu,\\
	0, & \left| x \right| > \mu,
	\end{array} \right. 
	\end{flalign}
	that is to say, if and only if $\mathbf{x}_{k} - \widehat{\mathbf{x}}_{k}$ uniform, and $I \left( \mathbf{x}_{k} - \widehat{\mathbf{x}}_{k}; \mathbf{x}_{0,\ldots,k-1} \right) = 0$.
\end{corollary}

It can be shown that \eqref{MD} reduces to the conclusions in \cite{FangMLSP19} when $\mathbf{x}_{k} - \widehat{\mathbf{x}}_{k}$ is further assumed to be with a compact set.

\subsection{Connections with the Estimation Counterparts to Fano's Equality}

We first present the following Corollary~\ref{Fano} which can be proved simply by replacing $\mathbf{x}_k$ and $\mathbf{y}_{0,\ldots,k-1}$ by $\mathbf{x}$ and $\mathbf{y}$ respectively in Theorem~\ref{MIMOprediction}.

\begin{corollary} \label{Fano}
	Consider a random variable $\mathbf{x} \in \mathbb{R}$ with side information $\mathbf{y} \in \mathbb{R}^k$.
	Then, it holds for any estimator $\widehat{\mathbf{x}} = g \left( \mathbf{y} \right)$ that  
	\begin{flalign} \label{Fano1}
	\left[ \mathbb{E} \left( \left| \mathbf{x} - \widehat{\mathbf{x}} \right|^{p} \right) \right]^{\frac{1}{p}}
	\geq \frac{2^{h \left( \mathbf{x} | \mathbf{y} \right)}}{2 \Gamma \left( \frac{p+1}{p} \right) \left( p \mathrm{e} \right)^{\frac{1}{p}}},
	\end{flalign}
	where equality holds if and only if $\mathbf{x} - \widehat{\mathbf{x}}$ is with probability density 
	\begin{flalign} 
	f_{\mathbf{x} - \widehat{\mathbf{x}}} \left( x \right)
	= \frac{ \mathrm{e}^{- \left| x \right|^{p} / \left( p \mu^{p} \right)} }{2 \Gamma \left( \frac{p+1}{p} \right) p^{\frac{1}{p}} \mu},
	\end{flalign}
	and $I \left( \mathbf{x} - \widehat{\mathbf{x}}; \mathbf{y} \right) = 0$. In addition, if the side information $\mathbf{y}$ is absent, it follows that
	\begin{flalign} \label{Fano2}
	\left[ \mathbb{E} \left( \left| \mathbf{x} - \widehat{\mathbf{x}} \right|^{p} \right) \right]^{\frac{1}{p}}
	\geq \frac{2^{h \left( \mathbf{x} \right)}}{2 \Gamma \left( \frac{p+1}{p} \right) \left( p \mathrm{e} \right)^{\frac{1}{p}}},
	\end{flalign}
	where equality holds if and only if $\mathbf{x} - \widehat{\mathbf{x}}$ is with probability density 
	\begin{flalign} 
	f_{\mathbf{x} - \widehat{\mathbf{x}}} \left( x \right)
	= \frac{ \mathrm{e}^{- \left| x \right|^{p} / \left( p \mu^{p} \right)} }{2 \Gamma \left( \frac{p+1}{p} \right) p^{\frac{1}{p}} \mu}.
	\end{flalign}
\end{corollary}

It can then be verified that when $p=2$, \eqref{Fano1} and \eqref{Fano2} reduce to the so-called estimation counterparts to Fano's inequality \cite{Cov:06}:
\begin{flalign}
\left[ \mathbb{E} \left( \left| \mathbf{x} - \widehat{\mathbf{x}} \right|^{2} \right) \right]^{\frac{1}{2}}
\geq \frac{2^{h \left( \mathbf{x} | \mathbf{y} \right)}}{\left( 2 \pi \mathrm{e} \right)^{\frac{1}{2}}};~\left[ \mathbb{E} \left( \left| \mathbf{x} - \widehat{\mathbf{x}} \right|^{2} \right) \right]^{\frac{1}{2}}
\geq \frac{2^{h \left( \mathbf{x} \right)}}{\left( 2 \pi \mathrm{e} \right)^{\frac{1}{2}}}.
\end{flalign}
On the other hand, for $p=1$ and $p=\infty$, \eqref{Fano1} and \eqref{Fano2} reduce respectively to
\begin{flalign} 
\mathbb{E} \left| \mathbf{x} - \widehat{\mathbf{x}} \right|
\geq \frac{2^{h \left( \mathbf{x} | \mathbf{y} \right)}}{2\mathrm{e}};~\mathbb{E} \left| \mathbf{x} - \widehat{\mathbf{x}} \right|
\geq \frac{2^{h \left( \mathbf{x} \right)}}{2\mathrm{e}},
\end{flalign}
and
\begin{flalign} 
\esssup_{ f_{\mathbf{x} - \widehat{\mathbf{x}}} \left( x \right) > 0} \left| \mathbf{x} - \widehat{\mathbf{x}} \right|
\geq \frac{2^{h \left( \mathbf{x} | \mathbf{y} \right)}}{2};~\esssup_{ f_{\mathbf{x} - \widehat{\mathbf{x}}} \left( x \right) > 0} \left| \mathbf{x} - \widehat{\mathbf{x}} \right|
\geq \frac{2^{h \left( \mathbf{x} \right)}}{2}.
\end{flalign}
In this sense, \eqref{Fano1} and \eqref{Fano2} may be viewed as a generalizations of the estimation counterparts to Fano's inequality.

%It is now also clear that
%\begin{flalign}
%&\det \mathrm{E} \left[ \left( \mathbf{x}_{k} - \widehat{\mathbf{x}}_{k} \right) Q \left( \mathbf{x}_{k} - \widehat{\mathbf{x}}_{k} \right)^T \right] \nonumber \\
%&~~~~ = \det Q \min_{f} \left| \mathrm{supp} \left( \mathbf{x}_{k} - \widehat{\mathbf{x}}_{k} \right) \right| \nonumber \\
%&~~~~ \geq \frac{\det Q}{\left( 2 \pi \mathrm{e} \right)^n} 2^{2 h \left( \mathbf{x}_k | \mathbf{x}_{0,\ldots,k-1} \right)},
%\end{flalign}
%where $Q \in \mathbb{R}^{n \times n}$ is a positive definite matrix. Herein, equality holds if and only if $\mathbf{x}_{k} - \widehat{\mathbf{x}}_{k}$ is uniform and $I \left( \mathbf{x}_{k} - \widehat{\mathbf{x}}_{k}; \mathbf{x}_{0,\ldots,k-1} \right) = 0$.

\subsection{Viewpoint of Entropic Innovations}

We next present an innovations' perspective \cite{FangITW19} to view the term $I \left( \mathbf{x}_{k} - \widehat{\mathbf{x}}_{k}; \mathbf{x}_{0,\ldots,k-1} \right)$. 

\begin{proposition} For $\widehat{\mathbf{x}}_{k} = g_{k} \left( \mathbf{x}_{0,\ldots,k-1} \right)$, it always holds that
	\begin{flalign}
	&I \left( \mathbf{x}_{k} - \widehat{\mathbf{x}}_{k}; \mathbf{x}_{0,\ldots,k-1} \right) \nonumber \\
	&~~~~ = I \left( \mathbf{x}_{k} - \widehat{\mathbf{x}}_{k} ; \mathbf{x}_{0} - \widehat{\mathbf{x}}_{0}, \ldots, \mathbf{x}_{k-1} - \widehat{\mathbf{x}}_{k-1} \right).
	\end{flalign}
\end{proposition}

Stated alternatively, the mutual information between the current innovation and the previous data points is equal to that between the current innovation and the previous innovations.
Accordingly, the condition that 
$
I \left( \mathbf{x}_{k} - \widehat{\mathbf{x}}_{k}; \mathbf{x}_{0,\ldots,k-1} \right) = 0
$
is equivalent to that 
\begin{flalign}
I \left( \mathbf{x}_{k} - \widehat{\mathbf{x}}_{k} ; \mathbf{y}_{0} - \widehat{\mathbf{y}}_{0}, \ldots, \mathbf{y}_{k-1} - \widehat{\mathbf{y}}_{k-1}, \mathbf{x}_{0,\ldots,k} \right) = 0,
\end{flalign}
which in turn means that the current innovation $\mathbf{x}_{k} - \widehat{\mathbf{x}}_{k} $ contains no information of the previous innovations. This is a key link that facilitates the subsequent analysis in the asymptotic case.

\begin{corollary} \label{MIMOasymp}
	Consider a stochastic process $\left\{ \mathbf{x}_{k} \right\}, \mathbf{x}_{k} \in \mathbb{R}$. Denote the $1$-step prediction of $\mathbf{x}_{k}$ by $\widehat{\mathbf{x}}_{k} = g_{k} \left( \mathbf{x}_{0,\ldots,k-1} \right)$.
	Then,  
	\begin{flalign} \label{MIMOasymp1}
	\liminf_{k\to \infty} \left[ \mathbb{E} \left( \left| \mathbf{x}_{k} - \widehat{\mathbf{x}}_{k} \right|^{p} \right) \right]^{\frac{1}{p}}
	\geq \liminf_{k\to \infty} \frac{2^{h \left( \mathbf{x}_k | \mathbf{x}_{0,\ldots,k-1} \right)}}{2 \Gamma \left( \frac{p+1}{p} \right) \left( p \mathrm{e} \right)^{\frac{1}{p}}},
	\end{flalign}
	where equality holds if $\left\{ \mathbf{x}_{k} - \widehat{\mathbf{x}}_{k} \right\}$ is asymptotically white and with probability density 
	\begin{flalign} \label{asydistribution}
	\lim_{k \to \infty} f_{\mathbf{x}_{k} - \widehat{\mathbf{x}}_{k}} \left( x \right)
	= \frac{ \mathrm{e}^{- \left| x \right|^{p} / \left( p \mu^{p} \right)} }{2 \Gamma \left( \frac{p+1}{p} \right) p^{\frac{1}{p}} \mu}.
	\end{flalign}
\end{corollary}

{\it Proof.}
It is known from Theorem~\ref{MIMOprediction} that
\begin{flalign} 
\left[ \mathbb{E} \left( \left| \mathbf{x}_{k} - \widehat{\mathbf{x}}_{k} \right|^{p} \right) \right]^{\frac{1}{p}}
\geq \frac{2^{h \left( \mathbf{x}_k | \mathbf{x}_{0,\ldots,k-1} \right)}}{2 \Gamma \left( \frac{p+1}{p} \right) \left( p \mathrm{e} \right)^{\frac{1}{p}}}, \nonumber
\end{flalign} 
where equality holds if and only if $\mathbf{x}_{k} - \widehat{\mathbf{x}}_{k}$ is with probability density \eqref{distribution} and $I \left( \mathbf{x}_{k} - \widehat{\mathbf{x}}_{k}; \mathbf{x}_{0,\ldots,k-1} \right) = 0$. This, by taking $\liminf_{k\to \infty}$ on its both sides, then leads to
\begin{flalign} 
\liminf_{k\to \infty} \left[ \mathbb{E} \left( \left| \mathbf{x}_{k} - \widehat{\mathbf{x}}_{k} \right|^{p} \right) \right]^{\frac{1}{p}}
\geq \liminf_{k\to \infty} \frac{2^{h \left( \mathbf{x}_k | \mathbf{x}_{0,\ldots,k-1} \right)}}{2 \Gamma \left( \frac{p+1}{p} \right) \left( p \mathrm{e} \right)^{\frac{1}{p}}}. \nonumber
\end{flalign}
Herein, equality holds if $\mathbf{x}_{k} - \widehat{\mathbf{x}}_{k}$ is with probability density \eqref{distribution} and 
\begin{flalign}
&I \left( \mathbf{x}_{k} - \widehat{\mathbf{x}}_{k}; \mathbf{x}_{0,\ldots,k-1} \right) \nonumber \\
&~~~~ = I \left( \mathbf{x}_{k} - \widehat{\mathbf{x}}_{k} ; \mathbf{y}_{0} - \widehat{\mathbf{y}}_{0}, \ldots, \mathbf{y}_{k-1} - \widehat{\mathbf{y}}_{k-1}, \mathbf{x}_{0,\ldots,k} \right)= 0, \nonumber
\end{flalign}
as $k\to \infty$. Since that
\begin{flalign}
I \left( \mathbf{x}_{k} - \widehat{\mathbf{x}}_{k} ; \mathbf{y}_{0} - \widehat{\mathbf{y}}_{0}, \ldots, \mathbf{y}_{k-1} - \widehat{\mathbf{y}}_{k-1}, \mathbf{x}_{0,\ldots,k} \right)= 0 \nonumber
\end{flalign}
as $k\to \infty$ is equivalent to that $\mathbf{x}_{k} - \widehat{\mathbf{x}}_{k}$ is asymptotically white, equality in \eqref{MIMOasymp1} holds if $\left\{ \mathbf{x}_{k} - \widehat{\mathbf{x}}_{k} \right\}$ is asymptotically white and with probability density \eqref{asydistribution}.
\hfill$\square$

Strictly speaking, herein “white” should be “independent (over time)”; in the rest of the paper, however, we will use “white”
to replace “independent” for simplicity, unless otherwise specified. 
%Note also that “independent” is equivalent to “uncorrelated” for Gaussian processes.

%Particularly, if $\left\{ \mathbf{x}_{k} \right\}$ is a Markov process, then \eqref{MIMOasymp1} becomes
%\begin{flalign}
%\liminf_{k\to \infty} \left| \mathrm{supp} \left( \mathbf{x}_{k} - \widehat{\mathbf{x}}_{k} \right) \right|
%\geq \liminf_{k\to \infty} 2^{ h \left( \mathbf{x}_k | \mathbf{x}_{k-1} \right)},
%\end{flalign}
%where equality holds if and only if $\left\{ \mathbf{x}_{k} - \widehat{\mathbf{x}}_{k} \right\}$ is asymptotically white uniform.

When the sequence to be predicted is asymptotically stationary, we arrive at the following result.

\begin{corollary} \label{uniform}
	Consider an asymptotically stationary stochastic process $\left\{ \mathbf{x}_{k} \right\}, \mathbf{x}_{k} \in \mathbb{R}$. Denote the $1$-step prediction of $\mathbf{x}_{k}$ by $\widehat{\mathbf{x}}_{k} = g_{k} \left( \mathbf{x}_{0,\ldots,k-1} \right)$.
	Then,
	\begin{flalign} 
	\liminf_{k\to \infty} \left[ \mathbb{E} \left( \left| \mathbf{x}_{k} - \widehat{\mathbf{x}}_{k} \right|^{p} \right) \right]^{\frac{1}{p}}
	\geq \frac{2^{h_{\infty} \left( \mathbf{x} \right)}}{2 \Gamma \left( \frac{p+1}{p} \right) \left( p \mathrm{e} \right)^{\frac{1}{p}}},
	%		= \frac{1}{2\pi}\int_{-\infty}^{\infty} \ln \left[ \frac{S_{\mathbf{z}} \left( \omega \right) + \sigma_{\mathbf{v}}^2}{\sigma_{\mathbf{v}}^2} \right] \mathrm{d} \omega.   \nonumber
	\end{flalign}
	where $h_{\infty} \left( \mathbf{x} \right)$ denotes the entropy rate of $\left\{ \mathbf{x}_{k} \right\}$. Herein, equality holds if $\left\{ \mathbf{x}_{k} - \widehat{\mathbf{x}}_{k} \right\}$ is asymptotically white and with probability density \eqref{asydistribution}.
\end{corollary}

Corollary~\ref{uniform} follows directly from Corollary~\ref{MIMOasymp} by noting that for asymptotically stationary processes $\left\{ \mathbf{x}_{k} \right\}$, we have \cite{Cov:06} 
\begin{flalign} 
\liminf_{k\to \infty} h \left( \mathbf{x}_k | \mathbf{x}_{0,\ldots,k-1} \right) = \lim_{k\to \infty} h \left( \mathbf{x}_k | \mathbf{x}_{0,\ldots,k-1} \right)  = h_{\infty} \left( \mathbf{x} \right). \nonumber
\end{flalign}

As a matter of fact, if $\left\{ \mathbf{x}_{k} - \overline{\mathbf{x}}_{k} \right\}$ is asymptotically white and with probability density \eqref{asydistribution}, then, noting also that $\left\{ \mathbf{x}_{k} \right\}$ is asymptotically stationary, it holds that
\begin{flalign} \label{equality}
\lim_{k\to \infty} \left[ \mathbb{E} \left( \left| \mathbf{x}_{k} - \widehat{\mathbf{x}}_{k} \right|^{p} \right) \right]^{\frac{1}{p}}
= \frac{2^{h_{\infty} \left( \mathbf{x} \right)}}{2 \Gamma \left( \frac{p+1}{p} \right) \left( p \mathrm{e} \right)^{\frac{1}{p}}}.
%		= \frac{1}{2\pi}\int_{-\infty}^{\infty} \ln \left[ \frac{S_{\mathbf{z}} \left( \omega \right) + \sigma_{\mathbf{v}}^2}{\sigma_{\mathbf{v}}^2} \right] \mathrm{d} \omega.   \nonumber
\end{flalign}
In addition, we can show that \eqref{equality} holds if and only if $\left\{ \mathbf{x}_{k} - \overline{\mathbf{x}}_{k} \right\}$ is asymptotically white and with probability density \eqref{asydistribution}; in other words, the necessary and sufficient condition for achieving the prediction bounds asymptotically is that the innovation is asymptotically white and with probability density \eqref{asydistribution}.

\section{$\mathcal{L}_{p}$ Bounds in Recursive Algorithms}

In this section, we investigate the fundamental limitations in recursive algorithms. In particular, we first present the following generic $\mathcal{L}_{p}$ bounds on the recursive differences.

\begin{theorem} \label{recursivetheorem}
	Consider a recursive algorithm given by 
	\begin{flalign} \label{recursivesystem}
	\mathbf{x}_{k+1} = \mathbf{x}_{k} + g_{k} \left( \mathbf{x}_{0,\ldots,k} \right) + \mathbf{n}_{k},
	\end{flalign} 
	where $\mathbf{x}_{k} \in \mathbb{R}$ denotes the recursive state, $\mathbf{n}_{k} \in \mathbb{R}$ denotes the noise, and $g_{k} \left( \mathbf{x}_{0,\ldots,k} \right) \in \mathbb{R}$.
	%\pause
	Then,  
	\begin{flalign} 
	\left[ \mathbb{E} \left( \left| \mathbf{x}_{k+1} - \mathbf{x}_{k} \right|^{p} \right) \right]^{\frac{1}{p}}
	\geq \frac{2^{h \left( \mathbf{n}_k | \mathbf{n}_{0,\ldots,k-1}, \mathbf{x}_{0} \right)}}{2 \Gamma \left( \frac{p+1}{p} \right) \left( p \mathrm{e} \right)^{\frac{1}{p}}},
	\end{flalign}
	where equality holds if and only if $\mathbf{x}_{k+1} - \mathbf{x}_{k}$ is with probability density 
	\begin{flalign} \label{recursivedistribution}
	f_{\mathbf{x}_{k+1} - \mathbf{x}_{k}} \left( x \right)
	= \frac{ \mathrm{e}^{- \left| x \right|^{p} / \left( p \mu^{p} \right)} }{2 \Gamma \left( \frac{p+1}{p} \right) p^{\frac{1}{p}} \mu},
	\end{flalign}
	and $I \left( \mathbf{x}_{k+1} - \mathbf{x}_{k}; \mathbf{n}_{0,\ldots,k-1}, \mathbf{x}_{0} \right) = 0$.
\end{theorem}

Before we prove Theorem~\ref{recursivetheorem}, we first prove the following proposition.

\begin{proposition} \label{function1}
For the recursive algorithm given in \eqref{recursivesystem}, it holds that $\mathbf{x}_{k}$ is eventually a function of $\mathbf{n}_{0,\ldots,k-1}$ and $\mathbf{x}_{0}$, i.e., 
\begin{flalign} \label{function}
\mathbf{x}_{k} = l_{k} \left( \mathbf{n}_{0,\ldots,k-1}, \mathbf{x}_{0} \right).
\end{flalign} 
\end{proposition}

{\it Proof.} To begin with, it is clear that when $k=0$, \eqref{recursivesystem} reduces to  
\begin{flalign}
\mathbf{x}_{1} = \mathbf{x}_{0} + g_{0} \left( \mathbf{x}_{0} \right) + \mathbf{n}_{0}, \nonumber
\end{flalign}
and thus it holds that 
\begin{flalign}
\mathbf{x}_{1} = l_{0} \left( \mathbf{n}_{0}, \mathbf{x}_{0} \right), \nonumber
\end{flalign}
that is, \eqref{function} holds for $k=0$. Next, when $k=1$, 
\eqref{recursivesystem} is given by 
\begin{flalign}
\mathbf{x}_{2} = \mathbf{x}_{1} + g_{1} \left( \mathbf{x}_{0}, \mathbf{x}_{1} \right) + \mathbf{n}_{1}. \nonumber
\end{flalign}
As such, since $\mathbf{x}_{1}$ is a function of $\mathbf{n}_{0}$ and $\mathbf{x}_{0}$, we have 
\begin{flalign}
\mathbf{x}_{2} = l_{0} \left( \mathbf{n}_{0}, \mathbf{x}_{0} \right) + g_{1} \left( \mathbf{x}_{0}, l_{0} \left( \mathbf{n}_{0}, \mathbf{x}_{0} \right) \right) + \mathbf{n}_{1}. \nonumber
\end{flalign} 
In other words, $\mathbf{x}_{2}$ is a function of $\mathbf{n}_{0,1}$ and $\mathbf{x}_{0}$, and thus \eqref{function} holds for $k=1$. We may then repeat this process and show that \eqref{function} holds for any $k \geq 0$.
\hfill$\square$

We next prove Theorem~\ref{recursivetheorem} based upon Proposition~\ref{function1}.

{\it Proof of Theorem~\ref{recursivetheorem}.}
It is known from Lemma~\ref{maximum} that 
\begin{flalign}
\left[ \mathbb{E} \left( \left| \mathbf{x}_{k+1} - \mathbf{x}_{k} \right|^{p} \right) \right]^{\frac{1}{p}}
\geq \frac{2^{h \left(  \mathbf{x}_{k+1} - \mathbf{x}_{k} \right)}}{2 \Gamma \left( \frac{p+1}{p} \right) \left( p \mathrm{e} \right)^{\frac{1}{p}}}, \nonumber
\end{flalign} 
where equality holds if and only if $\mathbf{x}_{k+1} - \mathbf{x}_{k}$ is with probability density \eqref{recursivedistribution}.
Meanwhile,
\begin{flalign} 
&h \left( \mathbf{x}_{k+1} - \mathbf{x}_{k} \right) \nonumber \\
& = h \left( \mathbf{x}_{k+1} - \mathbf{x}_{k} |  \mathbf{n}_{0,\ldots,k-1}, \mathbf{x}_{0} \right) + I \left( \mathbf{x}_{k+1} - \mathbf{x}_{k}; \mathbf{n}_{0,\ldots,k-1}, \mathbf{x}_{0} \right) \nonumber \\
& = h \left( g_{k} \left( \mathbf{x}_{0,\ldots,k} \right) + \mathbf{n}_{k} |  \mathbf{n}_{0,\ldots,k-1}, \mathbf{x}_{0} \right) \nonumber \\
&~~~~ + I \left( \mathbf{x}_{k+1} - \mathbf{x}_{k}; \mathbf{n}_{0,\ldots,k-1}, \mathbf{x}_{0} \right). \nonumber
\end{flalign}
Then, due to Proposition~\ref{function1}, $g_{k} \left( \mathbf{x}_{0,\ldots,k} \right)$ is a function of $\mathbf{n}_{0,\ldots,k-1}$ and $\mathbf{x}_{0}$. Hence,
\begin{flalign} 
h \left( g_{k} \left( \mathbf{x}_{0,\ldots,k} \right) + \mathbf{n}_{k} |  \mathbf{n}_{0,\ldots,k-1}, \mathbf{x}_{0} \right) = h \left( \mathbf{n}_{k} |  \mathbf{n}_{0,\ldots,k-1}, \mathbf{x}_{0} \right). \nonumber
\end{flalign}
As a result,
$
2^{ h \left( \mathbf{x}_{k+1} - \mathbf{x}_{k} \right)} 
\geq 2^{h \left( \mathbf{n}_k | \mathbf{n}_{0,\ldots,k-1}, \mathbf{x}_{0} \right)},
$
where equality holds if and only if $I \left( \mathbf{x}_{k+1} - \mathbf{x}_{k}; \mathbf{n}_{0,\ldots,k-1}, \mathbf{x}_{0} \right) = 0$.
Therefore,
\begin{flalign}
\left[ \mathbb{E} \left( \left| \mathbf{x}_{k+1} - \mathbf{x}_{k} \right|^{p} \right) \right]^{\frac{1}{p}}
\geq \frac{2^{h \left( \mathbf{n}_k | \mathbf{n}_{0,\ldots,k-1}, \mathbf{x}_{0} \right)}}{2 \Gamma \left( \frac{p+1}{p} \right) \left( p \mathrm{e} \right)^{\frac{1}{p}}}, \nonumber
\end{flalign}
where equality holds if and only if $\mathbf{x}_{k+1} - \mathbf{x}_{k}$ is with probability density \eqref{recursivedistribution} and $I \left( \mathbf{x}_{k+1} - \mathbf{x}_{k}; \mathbf{n}_{0,\ldots,k-1}, \mathbf{x}_{0} \right) = 0$.
\hfill$\square$

It is clear that herein the lower bounds are determined completely by the conditional entropy of the current noise $\mathbf{n}_{k}$ conditioned on the past noises $\mathbf{n}_{0,\ldots,k-1}$ and the initial state of the recursive algorithm.

Herein, if $\mathbf{x}_{0}$ is chosen deteministically, then 
\begin{flalign} 
\left[ \mathbb{E} \left( \left| \mathbf{x}_{k+1} - \mathbf{x}_{k} \right|^{p} \right) \right]^{\frac{1}{p}}
\geq \frac{2^{h \left( \mathbf{n}_k | \mathbf{n}_{0,\ldots,k-1} \right)}}{2 \Gamma \left( \frac{p+1}{p} \right) \left( p \mathrm{e} \right)^{\frac{1}{p}}},
\end{flalign}
where equality holds if and only if $\mathbf{x}_{k+1} - \mathbf{x}_{k}$ is with probability density \eqref{recursivedistribution} and $I \left( \mathbf{x}_{k+1} - \mathbf{x}_{k}; \mathbf{n}_{0,\ldots,k-1} \right) = 0$.

The next corollary examines the asymptotic case.

\begin{corollary} 
	Consider a recursive algorithm given by 
	\begin{flalign}
	\mathbf{x}_{k+1} = \mathbf{x}_{k} + g_{k} \left( \mathbf{x}_{0,\ldots,k} \right) + \mathbf{n}_{k},
	\end{flalign}
	Then,  
	\begin{flalign} \label{recursive}
	\liminf_{k \to \infty} \left[ \mathbb{E} \left( \left| \mathbf{x}_{k+1} - \mathbf{x}_{k} \right|^{p} \right) \right]^{\frac{1}{p}}
	\geq \liminf_{k \to \infty} \frac{2^{h \left( \mathbf{n}_k | \mathbf{n}_{0,\ldots,k-1}, \mathbf{x}_{0} \right)}}{2 \Gamma \left( \frac{p+1}{p} \right) \left( p \mathrm{e} \right)^{\frac{1}{p}}},
	\end{flalign}
	where equality holds if $\left\{ \mathbf{x}_{k+1} - \mathbf{x}_{k} \right\}$ is asymptotically with probability density 
	\begin{flalign} \label{recursivedistribution2}
	\liminf_{k \to \infty} f_{\mathbf{x}_{k+1} - \mathbf{x}_{k}} \left( x \right)
	= \frac{ \mathrm{e}^{- \left| x \right|^{p} / \left( p \mu^{p} \right)} }{2 \Gamma \left( \frac{p+1}{p} \right) p^{\frac{1}{p}} \mu},
	\end{flalign}
	and $\lim_{k \to \infty} I \left( \mathbf{x}_{k+1} - \mathbf{x}_{k}; \mathbf{n}_{0,\ldots,k-1}, \mathbf{x}_{0} \right) = 0$.
\end{corollary}

We now present an ``entropic innovations" (i.e., ``recursive differences" in this case) perspective to view the term $I \left( \mathbf{x}_{k+1} - \mathbf{x}_{k}; \mathbf{n}_{0,\ldots,k-1}, \mathbf{x}_{0} \right)$. 

\begin{proposition} For the recursive algorithm
	\begin{flalign}
	\mathbf{x}_{k+1} = \mathbf{x}_{k} + g_{k} \left( \mathbf{x}_{0,\ldots,k} \right) + \mathbf{n}_{k},
	\end{flalign}
	it always holds that
	\begin{flalign}
	&I \left( \mathbf{x}_{k+1} - \mathbf{x}_{k}; \mathbf{n}_{0,\ldots,k-1}, \mathbf{x}_{0} \right) \nonumber \\
	&~~~~ = I \left( \mathbf{x}_{k+1} - \mathbf{x}_{k} ; \mathbf{x}_{0}, \mathbf{x}_{1} - \mathbf{x}_{0}, \ldots, \mathbf{x}_{k} - \mathbf{x}_{k-1} \right).
	\end{flalign}
\end{proposition}

{\it Proof.} 
Since $\mathbf{x}_{k} = \mathbf{x}_{k-1} + g_{k-1} \left( \mathbf{x}_{0,\ldots,k-1} \right) + \mathbf{n}_{k-1}$, we have 
\begin{flalign} 
&I \left( \mathbf{x}_{k+1} - \mathbf{x}_{k} ; \mathbf{n}_{0,\ldots,k-1}, \mathbf{x}_{0} \right) \nonumber \\
& = I \left( \mathbf{x}_{k+1} - \mathbf{x}_{k} ; \mathbf{n}_{0,\ldots,k-2}, \mathbf{x}_{k} - \mathbf{x}_{k-1} - g_{k-1} \left( \mathbf{x}_{0,\ldots,k-1} \right), \mathbf{x}_{0} \right).\nonumber
\end{flalign}
On the other hand, due to Proposition~\ref{function1}, $g_{k-1} \left( \mathbf{x}_{0,\ldots,k-1} \right)$ is a function of $\mathbf{n}_{0,\ldots,k-2}$ and $\mathbf{x}_{0}$. As such,
\begin{flalign} 
&I \left( \mathbf{x}_{k+1} - \mathbf{x}_{k} ; \mathbf{n}_{0,\ldots,k-2}, \mathbf{x}_{k} - \mathbf{x}_{k-1} - g_{k-1} \left( \mathbf{x}_{0,\ldots,k-1} \right), \mathbf{x}_{0} \right) \nonumber \\
& = I \left( \mathbf{x}_{k+1} - \mathbf{x}_{k} ; \mathbf{n}_{0,\ldots,k-2}, \mathbf{x}_{k} - \mathbf{x}_{k-1}, \mathbf{x}_{0} \right)
.\nonumber
\end{flalign}
We may repeat the previous steps until we eventually arrive at
\begin{flalign} 
&I \left( \mathbf{x}_{k+1} - \mathbf{x}_{k} ; \mathbf{n}_{0,\ldots,k-2}, \mathbf{x}_{k} - \mathbf{x}_{k-1} - g_{k-1} \left( \mathbf{x}_{0,\ldots,k-1} \right), \mathbf{x}_{0} \right) \nonumber \\
& = I \left( \mathbf{x}_{k+1} - \mathbf{x}_{k} ; \mathbf{n}_{0,\ldots,k-2}, \mathbf{x}_{k} - \mathbf{x}_{k-1}, \mathbf{x}_{0} \right) \nonumber \\
& = \cdots = I \left( \mathbf{x}_{k+1} - \mathbf{x}_{k} ;  \mathbf{x}_{1} - \mathbf{x}_{0}, \ldots, \mathbf{x}_{k} - \mathbf{x}_{k-1}, \mathbf{x}_{0} \right)
,\nonumber
\end{flalign}
which completes the proof. \hfill$\square$

As such, from the viewpoint of ``recursive differences", the condition 
\begin{flalign} 
\lim_{k \to \infty} I \left( \mathbf{x}_{k+1} - \mathbf{x}_{k}; \mathbf{n}_{0,\ldots,k-1}, \mathbf{x}_{0} \right)  = 0 
\end{flalign}
is equivalent to that
\begin{flalign} 
\lim_{k \to \infty} I \left( \mathbf{x}_{k+1} - \mathbf{x}_{k} ; \mathbf{x}_{0}, \mathbf{x}_{1} - \mathbf{x}_{0}, \ldots, \mathbf{x}_{k} - \mathbf{x}_{k-1} \right) = 0.
\end{flalign}
That is to say, equality in \eqref{recursive} holds if $\left\{ \mathbf{x}_{k+1} - \mathbf{x}_{k} \right\}$ is asymptotically white and with distribution density \eqref{recursivedistribution2}.

On the other hand, it may be verified that when $\left\{ \mathbf{n}_{k} \right\}$ is asymptotically stationary and $\mathbf{x}_{0}$ is deterministic, \eqref{recursive} reduces to 
	\begin{flalign}
\liminf_{k \to \infty} \left[ \mathbb{E} \left( \left| \mathbf{x}_{k+1} - \mathbf{x}_{k} \right|^{p} \right) \right]^{\frac{1}{p}}
\geq \frac{2^{h_{\infty} \left( \mathbf{n} \right)}}{2 \Gamma \left( \frac{p+1}{p} \right) \left( p \mathrm{e} \right)^{\frac{1}{p}}}.
\end{flalign}

More generally, we can prove the following result.

\begin{theorem} \label{general}
	Consider a recursive algorithm given by 
	\begin{flalign}
	r_{k+1} \left( \mathbf{x}_{0,\ldots,k+1} \right) = g_{k} \left( \mathbf{x}_{0,\ldots,k} \right) + \mathbf{n}_{k},
	\end{flalign} 
	where $r_{k+1} \left( \mathbf{x}_{0,\ldots,k+1} \right), g_{k} \left( \mathbf{x}_{0,\ldots,k} \right) \in \mathbb{R}$, and $\mathbf{n}_{k} \in \mathbb{R}$ denotes the noise. Then,  
	\begin{flalign} 
	\left[ \mathbb{E} \left( \left| r_{k+1} \left( \mathbf{x}_{0,\ldots,k+1} \right) \right|^{p} \right) \right]^{\frac{1}{p}}
	\geq \frac{2^{h \left( \mathbf{n}_k | \mathbf{n}_{0,\ldots,k-1}, \mathbf{x}_{0} \right)}}{2 \Gamma \left( \frac{p+1}{p} \right) \left( p \mathrm{e} \right)^{\frac{1}{p}}},
	\end{flalign}
	where equality holds if and only if $r_{k+1} \left( \mathbf{x}_{0,\ldots,k+1} \right)$ is with probability density 
	\begin{flalign}
	f_{r_{k+1} \left( \mathbf{x}_{0,\ldots,k+1} \right)} \left( x \right)
	= \frac{ \mathrm{e}^{- \left| x \right|^{p} / \left( p \mu^{p} \right)} }{2 \Gamma \left( \frac{p+1}{p} \right) p^{\frac{1}{p}} \mu},
	\end{flalign}
	and $I \left( r_{k+1} \left( \mathbf{x}_{0,\ldots,k+1} \right); \mathbf{n}_{0,\ldots,k-1}, \mathbf{x}_{0} \right) = 0$.
\end{theorem}

Theorem~\ref{general} can be proved by following similar procedures to those in the proof of Theorem~\ref{recursivetheorem}, to be more specific, by replacing $\mathbf{x}_{k+1} - \mathbf{x}_{k}$ with more generally $r_{k+1} \left( \mathbf{x}_{0,\ldots,k+1} \right)$ therein.

In fact, when we let 
\begin{flalign}
r_{k+1} \left( \mathbf{x}_{0,\ldots,k+1} \right) = \mathbf{x}_{k+1} - \mathbf{x}_{k}
\end{flalign} 
in Theorem~\ref{general}, it reduces to Theorem~\ref{recursivetheorem}. In addition, we may analyze similarly, for instance, the case where
\begin{flalign}
r_{k+1} \left( \mathbf{x}_{0,\ldots,k+1} \right) = \mathbf{x}_{k+1},
\end{flalign} 
which corresponds to 
\begin{flalign}
\mathbf{x}_{k+1} = g_{k} \left( \mathbf{x}_{0,\ldots,k} \right) + \mathbf{n}_{k},
\end{flalign} 
as well as the case that 
\begin{flalign}
r_{k+1} \left( \mathbf{x}_{0,\ldots,k+1} \right) = \mathbf{x}_{k+1} - 2 \mathbf{x}_{k} + \mathbf{x}_{k-1},
\end{flalign} 
corresponding to 
\begin{flalign}
\mathbf{x}_{k+1} = 2 \mathbf{x}_{k} - \mathbf{x}_{k-1} + g_{k} \left( \mathbf{x}_{0,\ldots,k} \right) + \mathbf{n}_{k}.
\end{flalign}

Note that in the recursive algorithm bounds (as well as the previous sequential prediction bounds) obtained in this paper, the classes of algorithms (learning algorithms or optimization algorithms) that can be applied are not restricted. In other words, the bounds are valid for arbitrary algorithms in practical use. On the other hand, no specific restrictions have been imposed on the distributions of the data or noise in general.

The fundamental performance bounds feature baselines for performance assessment and evaluation of various machine learning or optimization algorithms, by providing theoretical bounds that are to be compared with the true performances. Such baselines may function as fundamental benchmarks that separate what is possible and what is impossible, and can thus be applied to indicate how much room is left for performance improvement in algorithm design, or to avoid infeasible design specifications in the first place, saving time to be spent on unnecessary parameter tuning work that is destined to be futile.

\section{Conclusion}

In this paper, we have derived fundamental $\mathcal{L}_{p}$ bounds in sequential prediction as well as in recursive algorithms using an information-theoretic approach. The fundamental bounds are applicable to any causal algorithms while the data or noise involved can be with arbitrary distributions. For future research, it will be interesting to investigate further implications of the bounds and the corresponding necessary and sufficient conditions to achieve them.

\bibliographystyle{IEEEtran}
\bibliography{references}

\end{document}